\documentclass[journal,twoside]{IEEEtran} 

\usepackage{cite}
\usepackage{amsmath,amssymb,amsfonts}
\usepackage{algorithmic}
\usepackage{hyperref}
\usepackage{graphicx}
\usepackage{xcolor}
\usepackage{tabularx}
\usepackage{textcomp}
\usepackage{booktabs}
\def\BibTeX{{\rm B\kern-.05em{\sc i\kern-.025em b}\kern-.08em
    T\kern-.1667em\lower.7ex\hbox{E}\kern-.125emX}}

\begin{document}

\title{REN: Anatomically-Informed Mixture-of-Experts for Interstitial Lung Disease Diagnosis}

\author{Alec K. Peltekian, Halil Ertugrul Aktas, Gorkem Durak, Kevin Grudzinski,\\
Bradford C. Bemiss, Carrie Richardson, Jane E. Dematte, G. R. Scott Budinger,\\
Anthony J. Esposito, Alexander Misharin, Alok Choudhary, Ankit Agrawal, and Ulas Bagci%
\thanks{A. K. Peltekian and A. Choudhary are with the Department of Computer Science, Northwestern University McCormick School of Engineering and Applied Science, Chicago, IL, United States. K. Grudzinski, B. C. Bemiss, J. E. Dematte, A. J. Esposito, G. R. S. Budinger, and A. Misharin are with the Division of Pulmonary and Critical Care Medicine, Northwestern University Feinberg School of Medicine, Chicago, IL, United States. C. Richardson is with the Division of Rheumatology, Northwestern University Feinberg School of Medicine, Chicago, IL, United States. G. R. S. Budinger and A. Misharin are also with the Simpson Querrey Lung Institute for Translational Science, Northwestern University Feinberg School of Medicine, Chicago, IL, United States. A. Choudhary and A. Agrawal are with the Department of Electrical and Computer Engineering, Northwestern University McCormick School of Engineering and Applied Science, Chicago, IL, United States. H. E. Aktas, G. Durak, and U. Bagci are with the Machine \& Hybrid Intelligence Lab, Department of Radiology, Northwestern University Feinberg School of Medicine, Chicago, IL, United States.}
\thanks{This research was supported in part through a generous gift from K. Querrey and L. Simpson. We acknowledge the Quest high-performance computing facility and the Genomics Compute Cluster. G.R.S.B. was supported by a Chicago Biomedical Consortium grant, Northwestern University Dixon Translational Science Award, Simpson Querrey Lung Institute for Translational Science, the NIH (grant nos. P01AG049665, P01HL154998, U54AG079754, R01HL147575, R01HL158139, R01HL147290, R21AG075423 and U19AI135964), and the Veterans Administration (award no. I01CX001777). A.V.M. was supported by the NIH (grant nos. U19AI135964, P01AG049665, P01HL154998, U19AI181102, R01HL153312, R01HL158139, R01ES034350 and R21AG075423). A.C. was supported by NSF (grant no. OAC-2331329). A.A. was supported by the NIH (grant nos. U19AI135964 and R01HL158138), Simpson Querrey Lung Institute for Translational Science, and NSF (grant no. OAC-2331329). A.J.E. was supported by the NIH (grant no. L30HL149048). U.B. acknowledges the following grant: R01-HL171376. The funders had no role in the study design, data collection and analysis, decision to publish, or preparation of the manuscript. Corresponding author: U. Bagci (e-mail: ulas.bagci@northwestern.edu).}
}

\maketitle

\begin{abstract}
Mixture-of-Experts (MoE) architectures achieve scalable learning by routing inputs to specialized subnetworks through conditional computation. However, conventional MoE designs assume homogeneous expert capability and domain-agnostic routing—assumptions that are fundamentally misaligned with medical imaging, where anatomical structure and regional disease heterogeneity govern pathological patterns. We introduce \textit{Regional Expert Networks (REN)}, the first anatomically-informed MoE framework for medical image classification. REN encodes anatomical priors by training seven specialized experts, each dedicated to a distinct lung lobe or bilateral lung combination, enabling precise modeling of region-specific pathological variation. Multi-modal gating mechanisms dynamically integrate radiomics biomarkers with deep learning (DL) features extracted by convolutional (CNN), Transformer (ViT), and state-space (Mamba) architectures to weight expert contributions at inference. Applied to interstitial lung disease (ILD) classification on a 597-patient, 1,898-scan longitudinal cohort, REN achieves consistently superior performance: the radiomics-guided ensemble attains an average AUC of \textbf{0.8646 $\pm$ 0.0467}, a \textbf{+12.5\%} improvement over the SwinUNETR single-model baseline (AUC 0.7685, $p=0.031$). Lower-lobe experts reach AUCs of 0.88-0.90, outperforming DL baselines (CNN: 0.76-0.79) and mirroring known patterns of basal ILD progression. Evaluated under rigorous patient-level cross-validation, REN demonstrates strong generalizability and clinical interpretability, establishing a scalable, anatomically-guided framework potentially extensible to other structured medical imaging tasks. Code is available on our GitHub \href{https://github.com/NUBagciLab/MoE-REN}{https://github.com/NUBagciLab/MoE-REN}.
\end{abstract}

\section{Introduction}
In machine learning, \textit{Mixture-of-experts (MoE)} architectures include multiple expert networks that focus on different aspects of complex data distributions. The success of MoE architectures stems from conditional computation; dynamically routing inputs to the most relevant experts while maintaining computational efficiency. However, existing MoE frameworks primarily operate under unconstrained optimization structures that assume homogeneous expert capabilities and domain-agnostic routing mechanisms. We argue that this approach is fundamentally misaligned with medical imaging. Anatomical structure, physiological constraints, and regional disease heterogeneity impose domain-specific specialization requirements that generic routing cannot capture effectively.

Unlike natural image classification, where semantic regions may be arbitrarily distributed, medical images exhibit well-defined anatomical structure where pathological patterns follow predictable regional distributions governed by underlying physiological processes~\cite{bagciPRL,jaus2024anatomy}. Traditional MoE systems cannot leverage these domain-specific priors, leading to suboptimal expert utilization and reduced interpretability, a critical requirement for clinical adoption~\cite{wu2025learning}.

Interstitial lung disease (ILD) diagnosis provides a concrete example of these challenges. ILD encompasses over 200 diverse pulmonary disorders. Accurate interpretation of high-resolution computed tomography (HRCT) scans is essential for distinguishing ILD subtypes and guiding clinical management~\cite{mei2023interstitial, soffer2022artificial}. Since each anatomical region may exhibit varying disease manifestations, severity levels, and progression patterns within the same patient, conventional global analysis approaches struggle to provide both accurate diagnosis and interpretable outputs. Current DL methods for ILD classification treat the entire lung as a homogeneous unit, potentially diluting region-specific pathological signals and providing limited interpretability for clinical decision-making~\cite{humphries2024deep}.

\begin{figure*}[!t]
\centering
\includegraphics[width=0.9\textwidth]{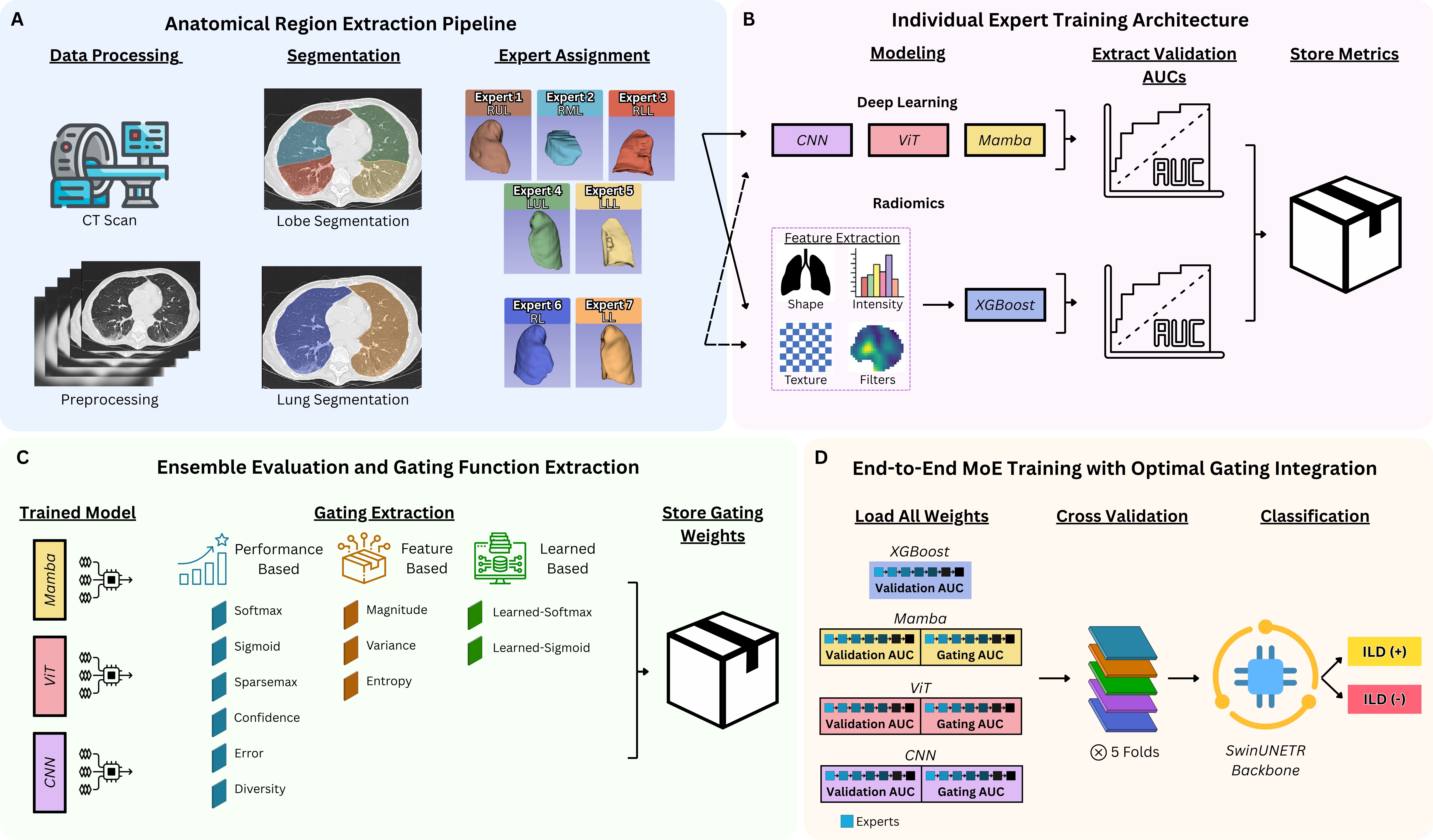}
\caption{Overview of the REN (Regional Expert Networks) framework. 
(A) Anatomical region extraction: preprocessing and lobe segmentation assign CT regions to seven experts (five lobes plus bilateral lungs). 
(B) Individual expert training: CNN, ViT, Mamba, and radiomics (XGBoost) experts are trained on masked inputs with validation AUCs recorded. 
(C) Gating function extraction: dynamic weighting strategies (performance-, feature-, and learned-based) are applied to expert outputs. 
(D) End-to-end MoE integration: expert weights and global SwinUNETR~\cite{hatamizadeh2021swin} features are fused for patient-level ILD classification.}
\label{fig:pipeline_overview}
\end{figure*}

To this end, we introduce REN (Regional Expert Networks), a novel anatomically-informed MoE framework that addresses the aforementioned limitations by embedding domain-specific constraints into expert specialization and routing mechanisms (Fig.~\ref{fig:pipeline_overview}). Unlike conventional MoE approaches where expert partitioning is learned implicitly, REN explicitly aligns expert specialization with anatomically defined lung regions, enabling interpretable regional modeling and structured expert routing tailored to ILD imaging patterns. Our novel approach demonstrates how anatomical priors can enhance MoE architectures. By training specialized experts for distinct anatomical regions and implementing multi-modal adaptive gating strategies, REN leverages inherent structural knowledge in medical applications while maintaining the scalability advantages of MoE systems. Through rigorous evaluation, we show that REN enables effective integration of both DL and traditional radiomics approaches. Particularly, radiomics-guided, newly designed gating strategies prove effective for capturing subtle pathological patterns in anatomically critical regions. \textbf{Our contributions} can be summarized as follows:

\begin{itemize}
\item \textbf{Novel Anatomically-Constrained MoE Architecture:} We introduce a new MoE framework specifically designed for medical imaging that incorporates anatomical structure as explicit constraints, addressing critical gaps in domain-specific expert routing. This is done by training seven lobe-specific experts that specialize in region-specific pathological pattern recognition through masked CT inputs.
\item \textbf{Multi-Modal Gating Mechanisms:} We develop new gating strategies that integrate both DL features and traditional radiomics biomarkers, to dynamically balance complementary information sources.
\item \textbf{Comprehensive Architecture Evaluation:} We systematically compare three state-of-the-art architectures—a custom 3D CNN~\cite{o2015introduction}, a 3D Vision Transformer adapted from Dosovitskiy et al.~\cite{dosovitskiy2020image}, and (Vision) Mamba~\cite{gu2023mamba} architecture—across all anatomical regions within the lung using rigorous patient-level cross-validation, establishing architecture-specific regional specialization insights.
\item \textbf{End-to-End Trainable Framework:} We implement a fully differentiable MoE system that enables joint optimization of expert networks and gating mechanisms while maintaining interpretability through expert contribution analysis.
\end{itemize}

\section{Related Work}
\subsection{Mixture of Experts in Deep Learning}  
MoE architectures leverage routing, expert diversity, and load balancing to enable scalable, efficient learning, as detailed in recent surveys~\cite{mu2025comprehensive}. While foundational for complex tasks, current MoE systems lack the domain-specific constraints critical for medical imaging, where anatomical structure guides disease patterns. A core challenge remains balancing expert specialization with computational cost. DeepSeek~\cite{dai2024deepseekmoe} advanced scalable MoE training through fine-grained expert partitioning and shared experts for generalization, whereas MoE++~\cite{jin2024moe} introduced zero-computation experts to enable efficient dynamic routing.

Extensions to vision and multi-modal settings have made MoE increasingly relevant to healthcare, while transformer-based architectures have also been explored for ILD lesion analysis and weakly supervised segmentation tasks~\cite{lai2024transformer}. MoE-LLaVA~\cite{lin2024moe} achieved sparse activation in vision-language models with only 3B active parameters, matching larger dense models via MoE-Tuning—a key step toward making MoE viable in resource-constrained clinical settings. Despite these advances, adapting MoE frameworks to incorporate anatomical priors and clinical semantics remains an open frontier for medical imaging.

Recent MoE research in medical imaging has also begun exploring domain-informed expert assignment and multi-modal integration. Wu et al.~\cite{wu2025learning} proposed heterogeneous tissue experts for gigapixel whole-slide images and Jiang and Shen~\cite{jiang2024m4oe} introduced M4oE, a foundation model for multi-modal segmentation using MoE routing. These efforts highlight growing interest in adapting MoE to biomedical applications.

Our work differs by introducing an anatomically-constrained MoE tailored specifically for systemic sclerosis-associated ILD. Unlike prior approaches that define experts by tissue type, modality, or distributional shifts, we explicitly align experts with anatomical regions—mirroring radiological practice and enabling interpretable, region-level contributions. Furthermore, we integrate handcrafted radiomics features with deep expert outputs, yielding hybrid gating strategies that couple domain priors with multi-modal signals. This anatomically- and clinically-grounded orientation distinguishes our study from broader MoE developments and addresses interpretability demands unique to medical imaging.

\subsection{Radiomics and Multi-Modal Integration}
Radiomics remains a powerful methodology in medical imaging, particularly when integrated with DL techniques. By extracting quantitative features describing texture, shape, and intensity patterns, radiomics provides complementary information to the hierarchical representations learned by deep networks~\cite{gillies2016radiomics, de2023role}. This synergy continues to demonstrate significant effectiveness in improving diagnostic accuracy through multi-modal fusion. Traditional machine learning classifiers like XGBoost~\cite{chen2016xgboost} maintain relevance for high-dimensional radiomics tasks due to their interpretability and capacity to model complex feature interactions. Nevertheless, current fusion strategies often overlook regional anatomical heterogeneity and differential diagnostic importance across tissue regions~\cite{bagciPET}. Anatomically-informed integration of radiomics with DL represents a promising yet underexplored frontier for enhancing both accuracy and clinical interpretability.

\subsection{Anatomical Specialization and Regional Analysis}
Integrating anatomical priors into DL models enhances performance and interpretability by emulating clinical radiology's regional examination practices~\cite{dalca2018anatomical,bagciPAN}. Regional analysis proves particularly effective for pathologies with anatomically specialized patterns~\cite{bagciregional}. Conventional approaches that rely on whole-organ analysis often assume anatomical homogeneity and risk overlooking localized variation in pathology~\cite{litjens2017survey,zhang2024diffboost}. In pulmonary imaging, for example, global feature extraction across entire lung volumes can obscure localized disease signals, reducing both sensitivity and interpretability~\cite{puttagunta2021medical,durak2025radiologic}. This underscores the clinical need for anatomically informed models that capture region-specific patterns and align outputs with radiological workflows~\cite{liu2021advances}.

Recent advances incorporate anatomical knowledge through: (i) Denoising autoencoders and specialized architectures~\cite{larrazabal2019anatomical}, (ii) Shape priors for segmentation accuracy~\cite{pham2019deep}, and (iii) Domain-informed constraints improving diagnostics~\cite{xie2021survey}. Brain tumor segmentation exemplifies significant gains from anatomical priors~\cite{wang2023improving}. This aligns naturally with MoE frameworks, where specialized subnetworks process distinct anatomical regions. The shift from generic to anatomically-informed expert routing represents a novel MoE application in medical imaging.

\subsection{Deep Learning for ILD Diagnosis}
Deep learning approaches for ILD analysis have grown substantially in recent years. Recent studies have explored CNN-based, multiple-instance-learning, radiomics-based, and transformer-based methods for tasks such as UIP prediction, lesion analysis, and disease pattern classification from high-resolution CT scans~\cite{humphries2024deep,lai2024transformer}. Comprehensive reviews highlight the rapid development of artificial intelligence systems for ILD detection and disease pattern analysis from CT imaging~\cite{soffer2022artificial}. Despite these advances, most existing approaches rely on global lung representations or task-specific pipelines rather than explicitly modeling anatomical regional specialization. REN instead introduces a mixture-of-experts framework that aligns expert specialization with anatomically defined lung regions.

\subsection{Summary of Gaps and Motivation}
Collectively, existing MoE frameworks lack the domain-specific constraints required for medical imaging; existing DL methods for ILD neglect regional anatomical structure; and existing radiomics–DL fusion strategies ignore regional heterogeneity. REN directly addresses all three limitations through an anatomically-constrained MoE design with multi-modal, region-aware gating.

\section{Methods}
Our proposed anatomically-informed mixture-of-experts framework for ILD diagnosis (classification) consists of a four-stage pipeline that progressively builds from individual expert training to gating function extraction and finally to end-to-end MoE model creation. Fig.~\ref{fig:pipeline_overview} represents REN in four stages: 
(A) anatomical region extraction through preprocessing and segmentation, (B) training of regional experts using CNN, ViT, Mamba, and radiomics models, (C) derivation of gating strategies to dynamically weight experts, and (D) end-to-end MoE integration with SwinUNETR for ILD classification. Global contextual features are extracted using SwinUNETR, a transformer-based medical imaging architecture that combines hierarchical Swin Transformer representations with UNet-style decoding for volumetric image analysis~\cite{hatamizadeh2021swin}.

\subsection{Dataset}
Our study utilized a retrospective dataset from the Northwestern Scleroderma Registry comprising 597 patients with 1,898 longitudinal chest CT scans acquired between 2001 and 2023. This study was approved by the Northwestern University Institutional Review Board. All participants provided informed consent for inclusion in the Northwestern Scleroderma Registry at the time of enrollment. The cohort included 489 (81.9\%) female patients with a mean age of 63.7 ± 12.7 years (range: 22.1-98.3 years). The study population was predominantly White (479 patients, 80.2\%), with 84 (14.1\%) Black, 20 (3.4\%) Asian, and 14 (2.3\%) other race patients. The dataset focused on systemic sclerosis-related conditions with the following disease subtype distribution: limited cutaneous systemic sclerosis (lcSSc): 284 (47.6\%), diffuse cutaneous systemic sclerosis (dcSSc): 245 (41.0\%), systemic sclerosis sine scleroderma (SSS): 28 (4.7\%), with other subtypes (6.7\%). Of the total cohort, 365 patients (61.1\%) had confirmed ILD (via multidisciplinary consensus), forming the positive class for our binary classification task. Patients contributed an average of 3.2 scans each, reflecting longitudinal monitoring patterns typical in systemic sclerosis care.

\subsection{Data Preprocessing and Evaluation Strategy}
\begin{figure*}[!t]
\centering
\includegraphics[width=1.6\columnwidth]{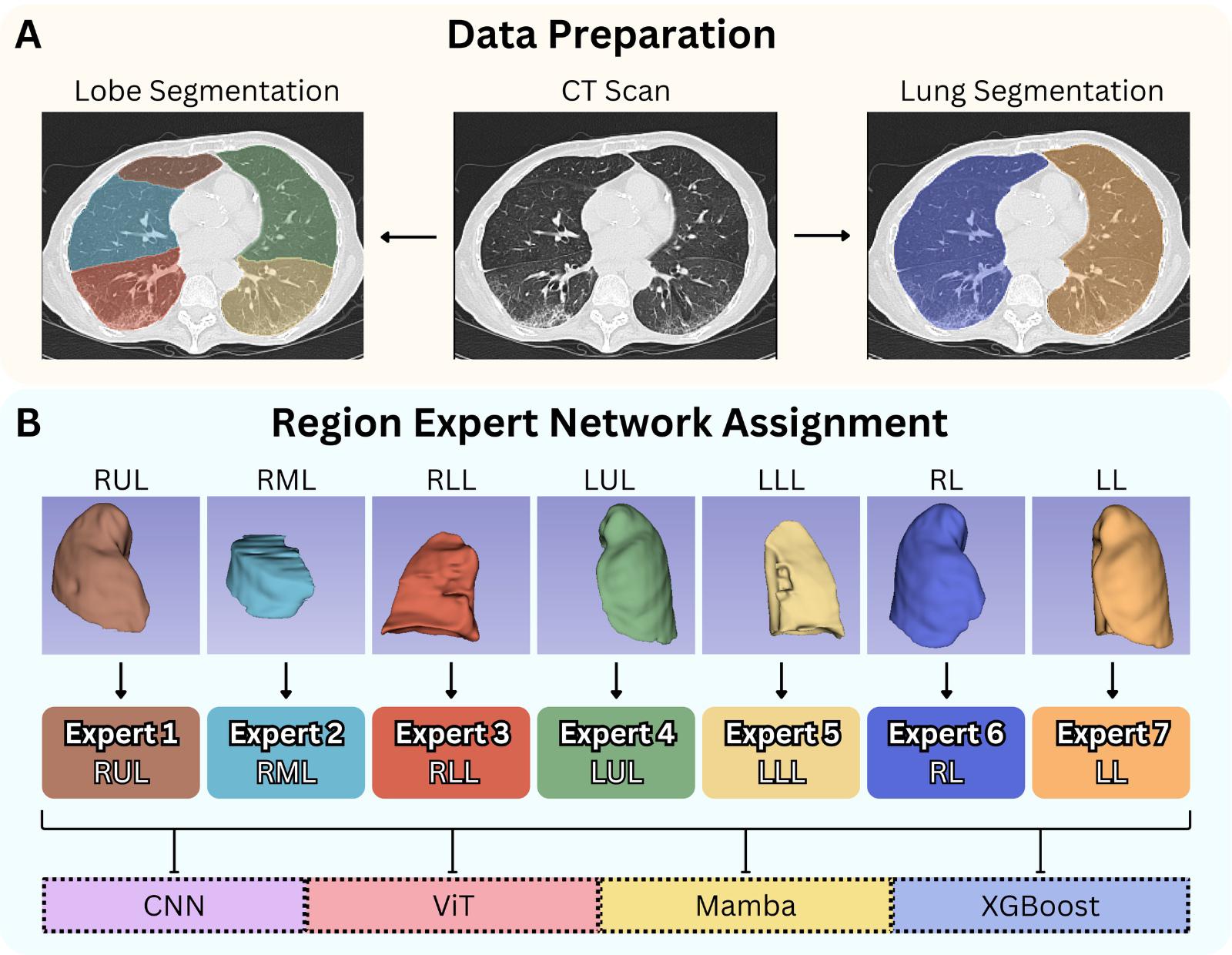}
\caption{Anatomical region extraction pipeline showing the process of generating masked inputs for each of the seven lung regions. The original CT scan is processed with lobe-specific segmentation masks to create region-focused inputs that enable expert specialization.}
\label{fig:region_extraction}
\end{figure*}

All CT scans, stored as NIFTI files, were processed using identical preprocessing pipelines, accompanied by corresponding lobe segmentation masks and binary classification labels. Lung and lobe segmentation were performed using the state-of-the-art \textit{lungmask} algorithm~\cite{hofmanninger2020automatic}, which provides robust automated segmentation of pulmonary lobes across diverse imaging protocols and pathological conditions. Our anatomical specialization approach focuses on seven distinct lung regions (automatically determined with \textit{lungmask} segmentation): the five individual lobes (Left Upper Lobe - LUL, Left Lower Lobe - LLL, Right Upper Lobe - RUL, Right Middle Lobe - RML, Right Lower Lobe - RLL) plus two combined regions (Left Lung combining lobes 1 and 2, Right Lung combining lobes 3, 4, and 5) (Fig.~\ref{fig:region_extraction}). For evaluation, we implement strict patient-level cross-validation to prevent data leakage, where multiple scans per patient could artificially inflate performance if distributed across training and evaluation sets.

Each fold assigns 80\% of patients to training and 20\% to holdout, where the 20\% holdout patients are unique to that fold. The holdout patients are then split 50/50 into validation (10\% of total patients) and test (10\% of total patients) sets (non-overlap). Across all 5 folds, every patient appears exactly once in the 20\% holdout position, ensuring 50\% of patients serve as validation data and 50\% as test data across folds, with no patient contamination between sets.

\subsubsection*{Feature-space visualization}
We qualitatively assessed regional separability by projecting radiomics feature vectors into two dimensions using t-SNE with standard settings and a fixed random seed. Embeddings were generated on the common set of scans across experts and are shown by anatomical region and by ILD status.

\subsection{Regional Expert Networks (RENs) Architecture}
Proposed REN modules specialize in distinct anatomical regions, each processing masked CT inputs corresponding to its assigned lung lobe or lung combination. This design ensures that learning is focused on region-specific pathological patterns rather than global averages. We implement several backbone variants: \textbf{CNN RENs} adopt a 3D convolutional architecture with three progressive blocks (24→48→96 filters, $3 \times 3 \times 3$ kernels), each followed by batch normalization, ReLU activation, and $2 \times 2 \times 2$ max pooling, before two fully connected layers (256→64 units) with dropout 0.5. \textbf{ViT RENs} partition volumes into $8 \times 8 \times 8$ patches, embed them into 384 dimensions, and process through 6 transformer blocks (8 attention heads, MLP ratio 4.0, GELU activation) with learnable positional encodings. \textbf{Mamba RENs} similarly use $8 \times 8 \times 8$ patches (256-dimensional embeddings) processed through 4 Mamba blocks ($d_{\text{state}}=64$, $d_{\text{conv}}=4$, expansion factor 2) with SiLU activation. This multi-architecture design allows comparative evaluation of regional specialization across convolutional, attention-based, and state-space modeling structures. REN should therefore be viewed as a clinically grounded adaptation of mixture-of-experts modeling that incorporates anatomical priors and multi-modal feature guidance rather than as a new generic MoE learning algorithm.

\subsection{Radiomics RENs}
For the five individual lung lobes and two lung regions, we extract comprehensive radiomics feature sets using PyRadiomics~\cite{van2017computational}. The radiomics pipeline computes first-order statistics (mean, variance, skewness, kurtosis), shape-based features (volume, surface area, sphericity, compactness), and texture features including Gray Level Co-occurrence Matrix (GLCM), Gray Level Run Length Matrix (GLRLM), Gray Level Size Zone Matrix (GLSZM), Gray Level Dependence Matrix (GLDM), and Neighbourhood Grey Tone Difference Matrix (NGTDM) features. This feature set captures complementary aspects of lung tissue characterization. First-order features quantify density and heterogeneity, shape features capture anatomical alterations, and texture features detect spatial and microstructural patterns associated with ILD-related pathophysiology. We used a 25 Hounsfield Unit bin width for intensity discretization to balance sensitivity to tissue differences while maintaining robustness to noise across different scanner protocols. This results in 107 quantitative features per lung lobe that capture subtle patterns potentially missed by DL approaches while providing clinically interpretable measurements. XGBoost classifiers are trained for each lobe and lung using these extracted features.

\subsection{Training Pipeline}
Our end-to-end approach follows four sequential stages: (Stage 1) anatomical preprocessing and lobe segmentation; (Stage 2) regional expert training; (Stage 3) ensemble and gating extraction; and (Stage 4) end-to-end MoE optimization.

\textbf{Stage 2 - Individual Regional Expert Network Training:} In Stage 2, we trained specialized RENs for seven anatomical lung regions using three DL architectures (CNN, ViT, Mamba) and one radiomics model (XGBoost). Each REN received a masked CT input, isolating the target region via element-wise multiplication with its anatomical mask. Deep learning models were trained with architecture-specific hyperparameters and medical imaging–oriented augmentations (rotation, noise, affine transforms). CNN and Mamba used a learning rate of $1 \times 10^{-4}$, batch size of 4;  ViT used a learning rate of $5 \times 10^{-5}$ and a batch size of 2 due to memory demands. All models were optimized with AdamW and regularized with dropout, early stopping, and label smoothing. This stage yielded 28 RENs per fold (7 regions × 4 models), with validation AUCs used in Stage 3 ensemble integration.

\textbf{Stage 3 – Staged Ensemble with Fixed Expert Weights.}
In Stage 3, regional experts are combined using gating weights derived from validation data, including validation-AUC–based weighting rules and learned gating networks. Crucially, once selected, these gating weights are \emph{fixed} and are not updated during subsequent inference or training. This staged design decouples expert learning from expert combination and preserves region-specific specialization.

\textbf{Stage 3A - Basic Ensemble Evaluation:}
Stage 3 includes two sub-steps. 
Stage 3A establishes a simple ensemble baseline by combining the outputs of the regional experts rather than relying on any single model. The motivation for using an ensemble is that different experts capture complementary signals from distinct lung regions, so aggregating them can reduce variance and improve overall stability. To formalize this, we compute a weighted prediction for each input, where the contribution of each expert is scaled by its validation AUC. This produces a single ensemble output that reflects both the predictions and the relative reliability of all experts.
For a given input sample $\mathbf{x}$ with corresponding lobe mask $\mathbf{m}$, each expert $E_k$ produces a prediction $\hat{y}_k(\mathbf{x})$, where $k \in \{1, 2, \ldots, K\}$ represents the seven anatomical regions with $K = 7$. 
Equation \eqref{eq:weighted_ensemble} defines the weighted ensemble prediction:
\begin{equation}
\hat{y}_{\text{weighted}}(\mathbf{x}) = \sum_{k=1}^{K} w_k \cdot \hat{y}_k(\mathbf{x}),
\label{eq:weighted_ensemble}
\end{equation}
where the normalized expert weights are computed as
\begin{equation}
w_k = \frac{\text{AUC}_k^{\text{val}}}{\sum_{j=1}^{K} \text{AUC}_j^{\text{val}}}.
\label{eq:val_auc_weight}
\end{equation}


\textbf{Stage 3B - Candidate Gating Strategy Evaluation:}
In this stage, the proposed gating formulations are treated as candidate expert-weighting rules (heuristics) rather than as a single unified learning algorithm. Each strategy defines a different mechanism for assigning weights to regional experts, enabling us to study how alternative weighting principles influence the combination of region-specific predictions.

We considered three classes of candidate strategies: (1) performance-based weighting rules derived from validation behavior, (2) feature-based weighting rules derived from intermediate representations, and (3) lightweight learned gating networks that produce input-dependent expert weights.

For each architecture and cross-validation fold, all candidate gating strategies were evaluated exclusively on the validation split. The strategy achieving the highest validation AUC was selected and then fixed for testing, ensuring no test-set leakage. Detailed mathematical formulations of the candidate weighting rules and learned gating architectures are provided in Appendix A.

\textbf{Definition of Ensemble Variants:}
For clarity and reproducibility, we explicitly define all ensemble configurations used throughout this study:
\begin{itemize}
\item \textbf{Radiomics-Guided}: Expert weights are derived exclusively from validation AUCs of radiomics (XGBoost) regional models and applied to deep expert outputs.
\item \textbf{Validation-Weighted}: Expert outputs are combined using fixed, global weights derived from normalized validation AUCs of the corresponding deep models.
\item \textbf{CNN / ViT / Mamba Gated MoE}: End-to-end MoE models in which expert weights are input-dependent and learned via gating networks that are jointly optimized with the backbone.
\end{itemize}

All reported results correspond to Tables~\ref{tab:final_comparison_5lobes}--\ref{tab:final_comparison_7regions}, with computational and ablation analyses reported in Tables~\ref{tab:training_complexity_5experts} and \ref{tab:gating_ablation}.



\textbf{Stage 4 - End-to-End MoE Architecture Training:}
Stage 4 initializes expert and gating parameters using the optimal weights identified in Stage 3 and then jointly optimizes the backbone, experts, and gating in an end-to-end manner. Unlike Stage 3, which preserves independent regional training, Stage 4 employs a unified architecture that processes the full lung volume while adaptively weighting anatomically constrained expert contributions.

SwinUNETR was selected as the Stage 4 backbone due to its strong and well-documented performance in volumetric medical imaging, its ability to balance global contextual modeling with local spatial fidelity~\cite{hatamizadeh2021swin}, and its widespread clinical adoption. The hierarchical Swin Transformer encoder captures long-range dependencies across lung regions, while the U-Net–style decoder preserves spatial structure, making it well suited for aggregating region-specific expert evidence into a unified representation. In our baseline experiments, SwinUNETR achieved higher full-lung AUC than alternative CNN-, ViT-, and Mamba-based backbones, motivating its use as the most competitive single-model reference. In addition, SwinUNETR benefits from large-scale medical imaging pretraining, which improves optimization stability and generalization in limited-data regimes. Importantly, REN is backbone-agnostic, and SwinUNETR is used here to provide a robust and reproducible reference for evaluating the proposed anatomically informed MoE formulation rather than to introduce architectural novelty.

\textbf{Architecture Design:} The proposed REN model processes the full CT volume $\mathbf{x}\in\mathbb{R}^{D\times H\times W}$ and lobe mask $\mathbf{m}\in\mathbb{R}^{D\times H\times W}$ using three main components. The SwinUNETR backbone extracts global features $\mathbf{F}_{\text{backbone}}=\text{SwinUNETR}(\mathbf{x})$, globally pooled to $\boldsymbol{\phi}_{\text{global}}\in\mathbb{R}^{384}$. Seven expert extractors process masked inputs $\mathbf{x}_k=\mathbf{x}\odot\mathbf{m}_k$ to produce features $\boldsymbol{\phi}_k\in\mathbb{R}^{64}$, with attention weights $\alpha_k$. Final expert weights are $g_k=w_k\alpha_k$, normalized such that $\sum_{k=1}^K g_k=1$. Weighted expert features $\boldsymbol{\phi}_{\text{expert}}=\sum_{k=1}^K g_k\boldsymbol{\phi}_k$ are concatenated with global features to form $\boldsymbol{\phi}_{\text{combined}}=[\boldsymbol{\phi}_{\text{global}};\boldsymbol{\phi}_{\text{expert}}]\in\mathbb{R}^{448}$, classified by an MLP as $y_{\text{pred}}=\text{MLP}(\boldsymbol{\phi}_{\text{combined}})$.

\textbf{Weight Initialization Strategies:} Stage 4 initializes and compares seven strategies: four based on validation AUC weighting from Stage 2 (radiomics, CNN, ViT, Mamba) and three based on the optimal learned gating from Stage 3 ($\mathbf{w}^*_{\text{CNN}}, \mathbf{w}^*_{\text{ViT}}, \mathbf{w}^*_{\text{Mamba}}$). For each fold, the selected initialization weights were normalized within the anatomical hierarchy before end-to-end optimization. Full formulations are provided in Appendix A.

\textbf{Training Configuration:} The end-to-end SwinUNETR MoE is trained with AdamW using differentiated learning rates ($10^{-5}$ for backbone, $10^{-4}$ for MoE) and cosine annealing scheduling. The total loss is
\begin{equation}
\mathcal{L}_{\text{total}} = \mathcal{L}_{\text{CE}} + \lambda_{\text{gating}}\mathcal{L}_{\text{gating}} + \lambda_{\text{weight}}\mathcal{L}_{\text{weight}} + \lambda_{\text{diversity}}\mathcal{L}_{\text{diversity}},
\label{eq:total_loss}
\end{equation}
with hyperparameters $\lambda_{\text{gating}}=0.005$, $\lambda_{\text{weight}}=0.005$, and $\lambda_{\text{diversity}}=0.01$ selected via validation.

\subsection{Scope of Anatomical Priors}
We operationalize anatomical knowledge as an inductive bias through region-constrained expert assignment, rather than as a formal probabilistic prior. Specifically, expert specialization is enforced through anatomically defined region decomposition and masked inputs, reflecting radiological practice.

\subsection{Summary of Methodological Innovations}
First, we introduce RENs, the first anatomically informed MoE architecture in medical imaging, specifically in lung diseases. Expert assignment is guided explicitly by lobe- and lung-level segmentation rather than unconstrained routing. This ensures that expert specialization mirrors radiological practice and directly encodes anatomical priors into the model.

Second, we propose a \textbf{novel multi-modal gating framework} that integrates three complementary strategies: (i) performance-based weighting derived from validation AUCs, (ii) feature-based weighting using intermediate representations, and (iii) lightweight learned gating networks. This design enables dynamic, input-adaptive expert selection while avoiding the interpretability and load-balancing limitations of conventional MoE routing.

Third, we incorporate \textbf{radiomics-guided expert models}, providing handcrafted, pathology-aware biomarkers that complement deep learning features. By embedding radiomics in both regional experts and gating mechanisms, our framework enhances interpretability and sensitivity to subtle, anatomically localized disease patterns.

\textbf{Role of Radiomics in REN:}
We acknowledge that radiomics–deep learning fusion has been explored in prior ILD and thoracic imaging studies~\cite{mei2023interstitial}. In our current study, our contribution lies in how radiomics is used within REN: radiomics features are not fused at the representation level, but serve as region-specific reliability signals that guide expert weighting within an anatomically constrained MoE framework. This design preserves modularity between handcrafted and learned representations while enabling region-aware interpretability through anatomically grounded expert weighting.

Collectively, these innovations constitute a domain-specific adaptation of MoE tailored for ILD classification. REN is not a simple application of existing MoE systems; it is a redesign that leverages anatomical priors, radiomics integration, and hybrid gating to produce clinically interpretable predictions.

\section{Experiments and Results}
We evaluated REN through comprehensive patient-level cross-validation experiments across three DL architectures and multiple gating strategies, using two anatomical configurations (five individual lobes; five lobes plus two bilateral regions). Table~\ref{tab:contribution_mapping} maps each claimed contribution to its corresponding experimental evidence.


\begin{table*}[t]
\centering
\caption{Mapping of REN contributions to experimental evaluations.}
\footnotesize
\begin{tabular}{p{4.5cm} p{7.5cm} p{2.5cm}}
\hline
\textbf{Contribution} & \textbf{Experimental Variant} & \textbf{Evidence} \\
\hline
Anatomical expert specialization & Regional CNN, ViT, and Mamba experts trained per lung region & Tables II, III \\
Radiomics-guided expert weighting & Radiomics-guided ensemble weighting & Tables II, III \\
Staged ensemble vs end-to-end MoE & Validation-weighted ensemble vs gated MoE & Tables II, III \\
Expert routing behavior & Routing diversity and usage metrics under regularization & Table V \\
Computational feasibility & Training time, memory, and parameter analysis & Table IV \\
\hline
\end{tabular}
\label{tab:contribution_mapping}
\end{table*}


\textbf{Model-Specific Hyperparameters:} All DL experts were trained using model-specific hyperparameters optimized for medical imaging tasks. CNN and Mamba experts used learning rate $1 \times 10^{-4}$, batch size 4, while ViT experts used learning rate $5 \times 10^{-5}$, batch size 2. All models used AdamW optimizer with weight decay $1 \times 10^{-5}$. Early stopping was applied with a patience of 15 epochs. Learning rate scheduling used ReduceLROnPlateau for CNNs and CosineAnnealingLR for Mamba and ViT models. Dropout rates were 0.5 for CNNs, 0.2 for Mamba and ViT. Label smoothing was 0.05 for CNN, 0.2 for ViT and Mamba.

\textbf{Computational Environment and Reproducibility:} All experiments were conducted on NVIDIA A100-SXM4-80GB GPUs, using CUDA 12.1 and driver version 525.85.12. The software environment consisted of Python 3.8.20, PyTorch 2.4.1+cu121, MONAI 1.3.2, NumPy 1.24.4, scikit-learn 1.3.2, Pandas 2.0.3, XGBoost 2.1.4, and pyradiomics v3.1.0. Mamba models used mamba-ssm 2.2.2 for state-space computations.

For reproducibility, all random number generators were initialized with seed=42 using a seeding function that sets Python's random module, NumPy, PyTorch CPU and GPU generators, and enables CUDA deterministic operations. Cross-validation splits used fold-specific seeds to ensure consistent patient partitioning across all experiments.

\subsection{Stage 2: Individual Expert Performance}
\begin{figure}[!t]
\centering
\includegraphics[width=0.9\columnwidth]{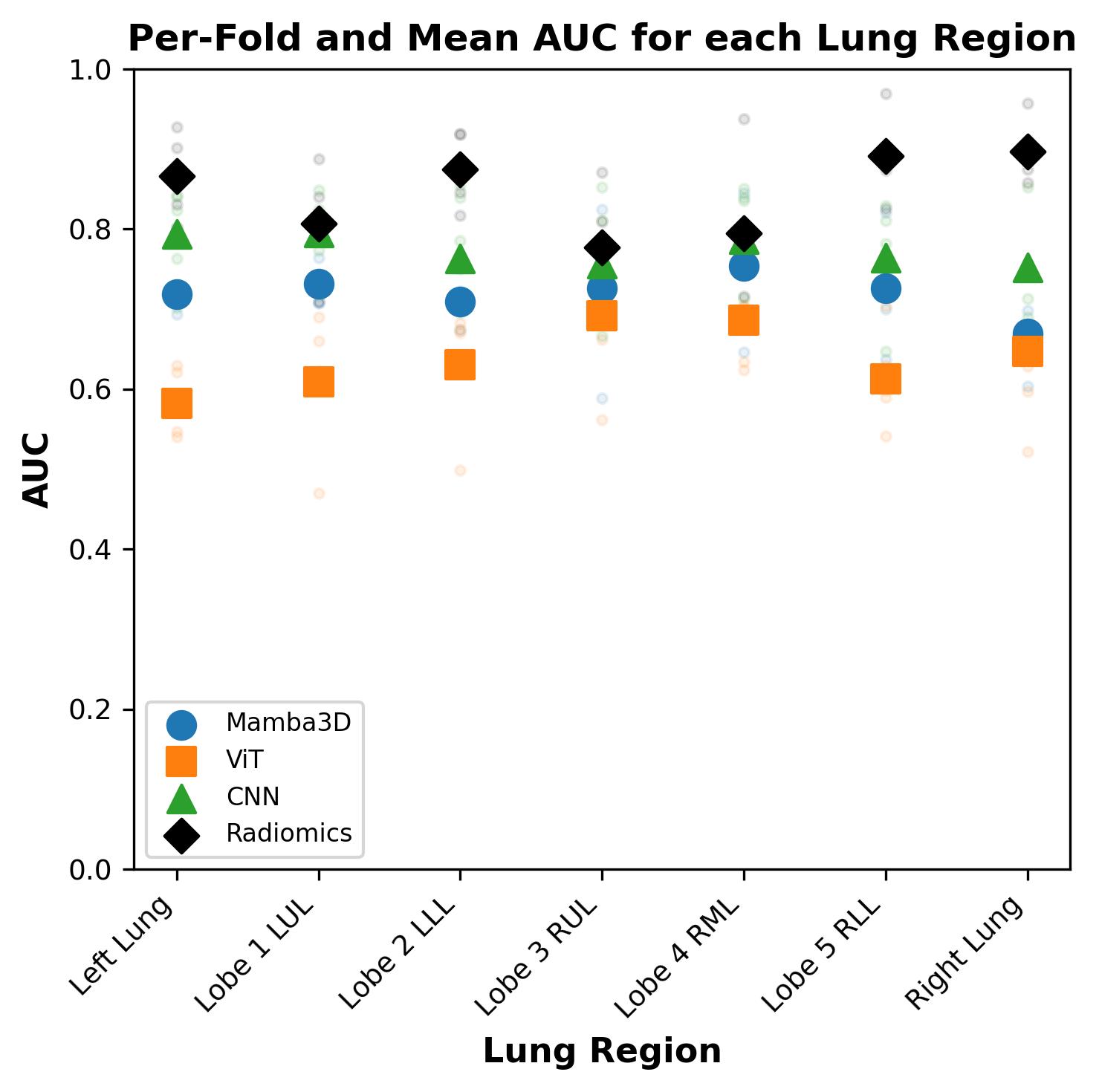}
\caption{Mean AUC per lung region across folds for each architecture. Radiomics experts achieved the highest regional performance, particularly in the lower lobes, while CNN followed closely behind. Mamba and ViT showed more variability across regions.}
\label{fig:auc_comparison_summary}
\end{figure}

\textbf{Architecture Performance Comparison:} Radiomics-based MoE models outperformed neural networks based MoEs in most region-fold comparisons (5 lobes × 5 folds = 25): 23/25 vs. Mamba, 24/25 vs. ViT, and 17/25 vs. CNN, with mean AUC differences of 0.117 ± 0.044 (95\% CI: [0.073, 0.162]), 0.193 ± 0.053 (95\% CI: [0.140, 0.246]), and 0.069 ± 0.046 (95\% CI: [0.022, 0.115]), respectively (Fig.~\ref{fig:auc_comparison_summary}). This performance advantage likely stems from radiomics’ domain-specific features (n=107), which encode texture, shape, and intensity, offering compact, pathology-aware representations and reducing overfitting risks on limited data. Among neural networks, CNNs showed the closest performance to radiomics, suggesting convolutional filters effectively model local lung structures. More advanced models, ViT and Mamba, originally optimized for non-medical domains, lagged behind—highlighting the need for adaptation when applying transformer-style models to clinical imaging~\cite{lai2024transformer}. Although the CNN expert contains substantially more parameters (42.6M) than the ViT (0.6M) and Mamba (0.7M) experts, all models were trained under identical data splits, preprocessing, and cross-validation conditions. Therefore, the observed performance differences reflect architectural inductive biases rather than differences in training protocol.

\textbf{Regional Specialization Patterns:} Model performance mirrored known ILD progression patterns, with highest AUCs in the left and right lower lobes (LLL/RLL), consistently exceeding 0.85 in the radiomics-based  MoE approach. This aligns with clinical evidence that ILD typically begins in the lung bases due to gravitational and mechanical stress. Radiomics was particularly sensitive to early fibrotic features—reticulation and ground-glass opacities—in these regions. In contrast, reduced performance in the upper lobes and right middle lobe (RML) across all methods reflects the later onset of ILD in these areas. These results suggest the models capture the true anatomical progression of ILD, rather than spurious spatial patterns.

\subsection{Stage 3: Learned and Gating Strategy Evaluation}
\textbf{Learned vs. Static Gating Performance:} Learned gating consistently outperforms static weighting across architectures within the gating strategy evaluation, improving model stability through adaptive expert selection. This validates dynamic gating as essential for medical MoE architectures, where input-dependent routing captures patient-specific anatomical variations while maintaining predictable behavior across diverse cases.

\subsection{Stage 4: End-to-End Integration Results} Tables~\ref{tab:final_comparison_5lobes} and \ref{tab:final_comparison_7regions} compare final test performance between weighted ensemble approaches, end-to-end MoE architectures, and baseline methods using different anatomical configurations.

\textbf{Recommended configuration:}
The highest diagnostic performance was achieved by the \textit{staged radiomics-guided REN ensemble using the five-lobe configuration} (average AUC 0.8646), which we therefore consider the primary accuracy-oriented configuration. The end-to-end gated MoE variants are included primarily to analyze expert routing and specialization under joint optimization rather than as the recommended deployment configuration. In traditional computer vision, joint optimization allows experts to dynamically partition the feature space. However, our ablation studies reveal that in the medical imaging problem we address, joint optimization causes partially overlapping expert representation that actively degrades the strict anatomical specialization (e.g., lobe-specific pathological patterns) required for ILD diagnosis. Hence, while end-to-end MoE is theoretically appealing, preserving independently trained experts via staged ensembling is fundamentally better suited for anatomically constrained medical tasks.

\begin{table}[!t]
\centering
\caption{Final Performance Using Five Individual Lung Lobes: Baseline vs. Validation-Weighted vs. End-to-End Gated MoE. Percentage change is relative to the SwinUNETR baseline (AUC 0.7685).}
\label{tab:final_comparison_5lobes}
\scriptsize
\begin{tabular}{lccc}
\toprule
\textbf{Method} & \textbf{AUC $\pm$ SD [95\% CI]} & \textbf{Change} & \textbf{p-value} \\
\midrule
\multicolumn{4}{c}{\textbf{Baseline}} \\
SwinUNETR (Baseline) & 0.7685 $\pm$ 0.0759 [0.674, 0.863] & -- & -- \\
CNN (Baseline) & 0.7584 $\pm$ 0.0919 [0.667, 0.850] & -- & -- \\
Mamba (Baseline) & 0.6775 $\pm$ 0.0454 [0.632, 0.723] & -- & -- \\
ViT (Baseline) & 0.6535 $\pm$ 0.0356 [0.618, 0.689] & -- & -- \\
\midrule
\multicolumn{4}{c}{\textbf{Validation-Weighted}} \\
CNN Weighted & 0.8033 $\pm$ 0.0472 [0.745, 0.862] & +4.5\% & 0.176 \\
ViT Weighted & 0.8015 $\pm$ 0.0453 [0.745, 0.858] & +4.3\% & 0.179 \\
Mamba Weighted & 0.7995 $\pm$ 0.0466 [0.742, 0.857] & +4.0\% & 0.213 \\
\textbf{Radiomics-Guided} & \textbf{0.8646 $\pm$ 0.0467 [0.806, 0.923]} & \textbf{+12.5\%} & \textbf{0.031} \\
\midrule
\multicolumn{4}{c}{\textbf{End-to-End Gated MoE}} \\
CNN Gated MoE & 0.7760 $\pm$ 0.0689 [0.691, 0.862] & +1.0\% & 0.819 \\
ViT Gated MoE & 0.8010 $\pm$ 0.0430 [0.748, 0.854] & +4.2\% & 0.166 \\
Mamba Gated MoE & 0.8164 $\pm$ 0.0393 [0.768, 0.865] & +6.2\% & 0.053 \\
\bottomrule
\end{tabular}
\end{table}

\begin{table}[!t]
\centering
\caption{Final Performance Comparison Using 5 Lobes + 2 Bilateral Lung Regions (7 Total): Baseline vs. Weighted MoE vs. Gated MoE. Percentage change is relative to the SwinUNETR baseline (AUC 0.7685).}
\label{tab:final_comparison_7regions}
\scriptsize
\begin{tabular}{lccc}
\toprule
\textbf{Method} & \textbf{AUC $\pm$ SD [95\% CI]} & \textbf{Change} & \textbf{p-value} \\
\midrule
\multicolumn{4}{c}{\textbf{Baseline}} \\
SwinUNETR (Baseline) & 0.7685 $\pm$ 0.0759 [0.674, 0.863] & -- & -- \\
CNN (Baseline) & 0.7584 $\pm$ 0.0919 [0.667, 0.850] & -- & -- \\
Mamba (Baseline) & 0.6775 $\pm$ 0.0454 [0.632, 0.723] & -- & -- \\
ViT (Baseline) & 0.6535 $\pm$ 0.0356 [0.618, 0.689] & -- & -- \\
\midrule
\multicolumn{4}{c}{\textbf{Validation-Weighted}} \\
CNN Weighted   & 0.7960 $\pm$ 0.0512 [0.732, 0.860] & +3.6\% & 0.382 \\
ViT Weighted   & 0.8021 $\pm$ 0.0577 [0.730, 0.874] & +4.4\% & 0.297 \\
Mamba Weighted & 0.7983 $\pm$ 0.0490 [0.737, 0.860] & +3.9\% & 0.351 \\
\textbf{Radiomics-Guided} & \textbf{0.8523 $\pm$ 0.0430 [0.796, 0.908]} & \textbf{+10.9\%} & \textbf{0.014} \\
\midrule
\multicolumn{4}{c}{\textbf{End-to-End Gated MoE}} \\
CNN Gated MoE   & 0.7778 $\pm$ 0.0545 [0.710, 0.846] & +1.2\% & 0.442 \\
ViT Gated MoE   & 0.7946 $\pm$ 0.0328 [0.754, 0.835] & +3.4\% & 0.242 \\
Mamba Gated MoE & 0.7950 $\pm$ 0.0486 [0.735, 0.855] & +3.4\% & 0.278 \\
\bottomrule
\end{tabular}
\end{table}



\begin{table}[!t]
\centering
\caption{Training and Inference Complexity Across Models}
\label{tab:training_complexity_5experts}
\scriptsize
\setlength{\tabcolsep}{2pt}
\renewcommand{\arraystretch}{1.05}

\begin{tabular}{lccc}
\toprule
\textbf{Model} & \textbf{Training Time} & \textbf{Peak Memory} & \textbf{Parameters} \\
 & \textbf{(GPU-hours)} & \textbf{(GB)} & \textbf{(Millions)} \\
\midrule
CNN Expert (single region)      & 0.82 & 1.94 & 42.6 \\
ViT Expert (single region)      & 0.82 & 1.66 & 0.6 \\
Mamba Expert (single region)    & 0.81 & 0.27 & 0.7 \\
SwinUNETR Baseline              & 0.91 & 12.98 & 62.2 \\
MoE Pipeline (CNN expert)       & 5.10 & 19.39 & 63.6 \\
MoE Pipeline (ViT expert)       & 4.98 & 19.39 & 63.6 \\
MoE Pipeline (Mamba expert)     & 5.04 & 19.39 & 63.6 \\
MoE Pipeline (Radiomics expert) & 5.07 & 19.39 & 63.6 \\
Gated MoE (end-to-end)          & 5.02 & 19.74 & 64.7 \\
\midrule
\multicolumn{4}{c}{\textbf{Inference Performance}} \\
\midrule
\textbf{Model} &
\textbf{Inference Time (ms)} &
\textbf{Samples/sec} &
\\
\midrule
{CNN / ViT / Mamba Experts} &
{46--48} &
{21+} &
\\
{Radiomics-Guided REN Ensemble} &
{62} &
{16} &
\\
\bottomrule
\end{tabular}
\end{table}

\begin{table}[!t]
\centering
\caption{Ablation Study of Gating Regularization Effects on Expert Behavior of all Experts}
\label{tab:gating_ablation}
\scriptsize
\setlength{\tabcolsep}{3pt}
\renewcommand{\arraystretch}{1.05}
\begin{tabular}{lcccc}
\toprule
\textbf{Configuration} &
\textbf{Mean Active} &
\textbf{Usage} &
\textbf{Mean} &
\textbf{Cosine} \\
&
\textbf{Experts} &
\textbf{Variance} &
\textbf{Entropy} &
\textbf{Similarity} \\
\midrule
Baseline (Full Loss)            & 3.10 & 237.2 & 1.43 & 0.49 \\
\midrule
No Regularization               & 2.25 & 920.2 & 0.85 & 0.83 \\
Remove Entropy                  & 2.58 & 620.2 & 1.00 & 0.41 \\
Remove Load Balancing           & 3.01 & 290.3 & 1.35 & 0.53 \\
Remove Diversity                & 3.34 & 183.2 & 1.49 & 0.82 \\
\midrule
Entropy Weight = 0.001          & 2.80 & 532.8 & 1.12 & 0.44 \\
Entropy Weight = 0.002          & 2.69 & 673.4 & 1.07 & 0.52 \\
Entropy Weight = 0.01           & 3.76 & 97.5  & 1.68 & 0.54 \\
Entropy Weight = 0.02           & 5.12 & 60.7  & 1.79 & 0.60 \\
\midrule
Load Balance Weight = 0.001     & 3.36 & 238.0 & 1.44 & 0.52 \\
Load Balance Weight = 0.005     & 3.49 & 142.2 & 1.58 & 0.48 \\
Load Balance Weight = 0.02      & 3.48 & 226.1 & 1.45 & 0.57 \\
Load Balance Weight = 0.05      & 3.11 & 305.4 & 1.38 & 0.61 \\
\midrule
Diversity Weight = 0.001        & 3.03 & 223.6 & 1.38 & 0.83 \\
Diversity Weight = 0.005        & 3.05 & 256.7 & 1.40 & 0.60 \\
Diversity Weight = 0.02         & 2.61 & 572.2 & 1.21 & 0.38 \\
Diversity Weight = 0.05         & 2.52 & 721.3 & 1.12 & 0.29 \\
\bottomrule
\end{tabular}
\end{table}

\textbf{Anatomical Configuration Comparison:} We evaluated two anatomical expert configurations: (1) five individual lung lobes only, and (2) seven regions including five individual lobes plus bilateral lung combinations. The five-lobe configuration achieved superior performance, with the radiomics-guided ensemble reaching an average AUC of 0.8646 compared to 0.8523 for the seven-region configuration. This 1.4\% improvement suggests that including bilateral lung experts may introduce redundancy that dilutes the specialized knowledge of individual lobe experts.

\textbf{Primary Performance and Computational Burden:} The radiomics-guided ensemble achieved the highest performance using the five-lobe configuration (average AUC 0.8646). This represents a statistically significant 12.5\% improvement over baseline SwinUNETR ($p = 0.031$, paired t-test across CV folds). This demonstrates the complementary value of engineered features when combined with anatomically informed expert weighting. All MoE variants meet typical clinical deployment requirements. Inference times are 46–48 ms per patient (21+ samples/s) for DL approaches and 62 ms for radiomics-guided ensembles. Model sizes (237–247 MB; 62–65 M parameters) remain manageable for standard clinical hardware. These speeds far exceed human interpretation time, enabling real-time diagnostic support. Importantly, weighted ensembles consistently outperform end-to-end gated counterparts. This validates our training strategy: staged ensembling preserves specialized anatomical knowledge that joint optimization obscures. The modest computational overhead of radiomics integration ($\approx 16$ ms) is negligible relative to its 12.5\% performance gain. This establishes the hybrid approach as both computationally feasible and diagnostically superior for clinical deployment.

Detailed training/inference time, memory usage, and parameter counts for all expert and MoE variants are reported in Table~\ref{tab:training_complexity_5experts}. For clarity, the reported complexity for CNN, ViT, and Mamba models corresponds to \textit{per-expert training}, while the MoE pipeline rows represent the computational cost of the \textit{full REN ensemble}, including all regional experts and gating components. Although staged MoE pipelines incur higher total training cost due to independent expert training, peak memory usage remains below 20\,GB and per-model training time is comparable to the SwinUNETR baseline, supporting practical clinical deployment.

\textbf{Clinical Translation and Significance:} The 12.5\% improvement could translate to better identification of early-stage ILD cases, where timely diagnosis is critical for patient outcomes. The statistical significance of only the radiomics-guided approach suggests that while other methods show numerical improvements, they may not provide reliable enough enhancement for clinical implementation.

\textbf{Ablation of Gating Regularization Terms:}
Table~\ref{tab:gating_ablation} analyzes the effect of individual gating regularizers on expert utilization and redundancy. Removing all regularization leads to severe expert collapse, characterized by high usage variance and strong feature similarity between experts. Entropy regularization controls the number of active experts and prevents overconfident routing, while load-balancing regularization reduces expert usage variance and promotes equitable expert participation. Diversity regularization plays a critical role in reducing redundancy, as removing it substantially increases cosine similarity between expert representations. Together, these components stabilize expert behavior, balancing specialization, coverage, and interpretability.

\begin{figure*}[t]
\includegraphics[width=\textwidth]{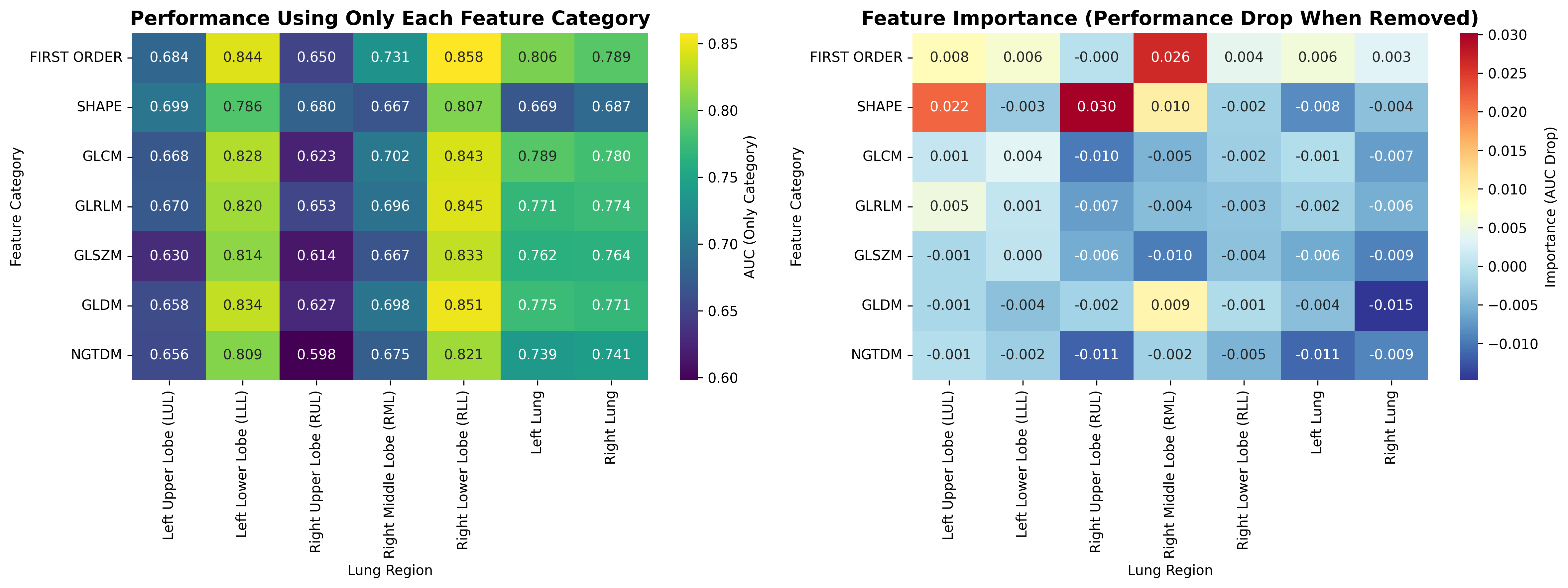}
\caption{Radiomics feature category ablation analysis showing (left) performance using individual feature categories across all lung regions and (right) feature importance measured by AUC drop when each category is removed. Texture features demonstrate particular strength in lower lobe regions, with GLCM achieving 0.828 AUC in Left Lower Lobe and GLRLM reaching 0.845 in Right Lower Lobe.}
\label{fig:ablation}
\end{figure*}

\textbf{Radiomics Analysis:} Comprehensive ablation studies reveal texture features (GLCM, GLRLM) drive performance in lower lobes where ILD initiates (Fig.~\ref{fig:ablation}). The embeddings corroborate the quantitative results: texture families (GLCM/GLRLM) yield tight, separable lower-lobe clusters, while upper-lobe geometry remains fragmented, explaining regional AUC differences. Lower lobe clustering patterns align with clinical knowledge of disease progression, while the approach's consistency across cross-validation folds ensures reliability across diverse clinical scenarios—critical for real-world deployment. The 7.6-8.1\% improvement over individual ensembles demonstrates tangible workflow benefits. This hybrid approach combines radiomics interpretability with DL's pattern recognition, providing both diagnostic accuracy and explainable insights. Regional performance breakdowns enable targeted radiologist attention and support differential diagnosis based on anatomical distribution patterns, aligning with natural radiological evaluation practices and facilitating clinical integration.

\section{Discussion and Concluding Remarks}
REN demonstrates a principled and effective approach to anatomically-informed MoE architectures in medical imaging. By aligning expert specialization with lobe-level anatomy and integrating radiomics biomarkers via multi-modal gating, REN consistently outperforms both single-model baselines and unconstrained ensemble approaches for ILD classification. The 12.5\% AUC gain over the SwinUNETR baseline is statistically significant and clinically meaningful, as improved sensitivity at this margin could translate directly to earlier identification of ILD—where timely diagnosis is critical for slowing disease progression.

Three key insights emerge from the experimental results. First, cross-architecture comparison reveals region-specific performance profiles: CNN and radiomics experts excel in lower lobes where ILD initiates, while transformer-based architectures offer complementary sensitivity in regions with more diffuse involvement. This suggests that heterogeneous expert assignment—selecting the architecture best suited to each anatomical region—may yield further gains over uniform architectural deployment. Second, staged ensembling consistently and substantially outperforms end-to-end joint optimization. When experts are jointly trained, shared gradients erode the anatomical specialization that makes each expert informative; the ablation data in Table V make this collapse explicit. Preserving independently trained regional experts and combining them via fixed gating is the correct design choice for anatomically-constrained tasks. Third, the performance advantage of radiomics-guided weighting arises not from radiomics features alone, but from their synergy with lobe-specific expert assignment. Whole-lung radiomics models perform substantially worse than lobe-specific radiomics experts, confirming that anatomical decomposition is a prerequisite for effective handcrafted feature utilization.

Anatomical grounding also confers direct interpretability benefits. Gating weights provide per-patient, per-region evidence maps that enable radiologists to identify which lung lobes drove the model's prediction—a form of spatial attribution that aligns naturally with existing reporting workflows. Unlike saliency maps or attention visualizations that require post-hoc analysis, REN's lobe-level contributions are intrinsic to the model architecture and operationally interpretable without additional computation.

Dense expert activation is a deliberate design choice that distinguishes REN from sparsely-activated MoE architectures. Clinical radiology requires systematic assessment of all anatomically relevant regions; enforcing hard sparsity would suppress informationally relevant experts for computational savings that are unnecessary given REN's already-feasible 62 ms inference latency. Soft sparsity through gating down-weighting uninformative regions per patient achieves an appropriate balance between anatomical completeness and computational efficiency.

For clinical deployment, REN is intended as a decision-support module rather than an autonomous diagnostic system. It processes routine chest CT scans, performs automated lobe segmentation and regional analysis, and outputs a global ILD confidence label with calibrated per-lobe expert contributions. Practical deployment requires robust handling of protocol variability and common confounders (infection, edema, motion, emphysema), quality-control checks, and a human-in-the-loop workflow. Before clinical use, multi-center external validation and prospective evaluation are necessary, along with governance aligned with privacy and regulatory standards.

Model compression for deployment is tractable due to REN's staged design. Knowledge distillation—training a compact student network using REN ensemble soft predictions and region-level intermediate features—is the most natural compression strategy. Expert pruning (removing experts with consistently low gating weights), top-k regional routing, and post-training quantization offer further options for latency and memory reduction without sacrificing diagnostic performance.

\textbf{Limitations:} Several limitations constrain the present findings. The dataset originates from a single institution and focuses exclusively on systemic sclerosis-related ILD, which may limit generalizability to other ILD subtypes and patient populations. The 2001–2023 temporal range introduces variability in imaging protocols and scanner hardware; while this serves as a natural domain-shift stress test (1,898 scans across 11 affiliated hospitals over 22 years), it is not a substitute for prospective multi-center validation. Reliance on automated lungmask segmentation may propagate boundary errors throughout the pipeline; future work will evaluate sensitivity to segmentation quality through controlled perturbation experiments. The current binary ILD classification does not capture heterogeneity across subtypes, and fixed anatomical expert assignment may limit adaptability to disease-specific spatial patterns (e.g., mid-upper lung involvement in hypersensitivity pneumonitis). The ensemble's total computational cost—from training and deploying seven independent regional networks—may also present scalability challenges in resource-constrained settings.

Although cross-institutional data sharing currently precludes additional external validation, ongoing multi-institutional collaborations are planned to evaluate REN across heterogeneous scanners, patient demographics, and ILD etiologies under domain shift.

\textbf{Future Directions:} We identify four priority directions for future work: (1) multi-institutional validation across demographics, scanner types, and ILD etiologies; (2) expansion to multi-class ILD subtype classification; (3) knowledge distillation strategies that compress the ensemble while preserving the observed 12.5\% performance gain; and (4) adaptive expert assignment driven by pathological distribution rather than fixed anatomical regions. Incorporating graph-based MoE formulations~\cite{wang2023graph} could enable relational routing that captures anatomical adjacency and disease-spread dynamics across lobes. Extension to vision-language domains—integrating imaging with radiology reports via multimodal MoE architectures such as MedMoE~\cite{chopra2025medmoe}, represents a natural next step toward clinically actionable AI that reasons jointly over images and text.

\appendices
\section{Methods and Mathematical Formulations}

This appendix provides detailed mathematical formulations and architectural specifications for the gating strategies and expert-weighting mechanisms used in REN. These formulations are included to ensure full reproducibility while keeping the main manuscript focused on the core architectural design and experimental evaluation.

\subsection{Candidate Gating Strategy Formulations}

\subsubsection{Enhanced AUC Normalization}
Various normalization schemes, such as softmax, sigmoid, and sparsemax are applied to validation AUC scores in \eqref{eq:auc_softmax}–\eqref{eq:auc_sparsemax},
\begin{align}
g_k^{\text{val-auc-softmax}} &= \frac{\exp(\text{AUC}_k^{\text{val}})}{\sum_{j=1}^{K} \exp(\text{AUC}_j^{\text{val}})}, \label{eq:auc_softmax} \\
g_k^{\text{val-auc-sigmoid}} &= \frac{\sigma(\text{AUC}_k^{\text{val}})}{\sum_{j=1}^{K} \sigma(\text{AUC}_j^{\text{val}})}, \label{eq:auc_sigmoid} \\
g_k^{\text{val-auc-sparsemax}} &= \max(\text{AUC}_k^{\text{val}} - \tau, 0), \label{eq:auc_sparsemax}
\end{align}
where $\sigma(x) = 1/(1+e^{-x})$ and $\tau$ ensures unity sum for sparsemax. Softmax emphasizes small differences between AUC scores by producing dense probability distributions, sigmoid compresses the range of values to create smoother weightings, and sparsemax projects onto the simplex with exact zeros to encourage sparsity. These approaches differ in how strongly they amplify, smooth, or prune expert contributions.

In sparsemax normalization, the threshold $\tau$ is computed as the unique value that ensures $\sum_k \max(\text{AUC}_k^{\text{val}} - \tau, 0) = 1$, following the standard sparsemax formulation.

\subsubsection{Dynamic Performance Gating}
Sample-dependent weights based on prediction characteristics are defined in \eqref{eq:confidence}–\eqref{eq:diversity},
\begin{align}
g_k^{\text{confidence}}(\mathbf{x}) &= \frac{\exp\!\left(2|\hat{y}_k(\mathbf{x})-0.5|\right)}{\sum_{j=1}^{K} \exp\!\left(2|\hat{y}_j(\mathbf{x})-0.5|\right)}, \label{eq:confidence} \\
g_k^{\text{error}}(\mathbf{x}) &= \frac{\exp\!\left(-|\hat{y}_k(\mathbf{x})-y|\right)}{\sum_{j=1}^{K} \exp\!\left(-|\hat{y}_j(\mathbf{x})-y|\right)}, \label{eq:error} \\
g_k^{\text{diversity}} &= \frac{\exp(1-\bar{\rho}_k)}{\sum_{j=1}^{K} \exp(1-\bar{\rho}_j)}, \label{eq:diversity}
\end{align}
where $\bar{\rho}_k$ represents the average correlation between expert $k$ and all others. Confidence, error, and diversity each address complementary aspects of expert reliability. Confidence weighting gives more influence to experts that make decisive predictions far from uncertainty, error weighting prioritizes experts that consistently align with the ground truth, and diversity weighting favors experts that provide outputs less correlated with the others.

\subsubsection{Statistical Feature Analysis}
Feature-vector-based weighting strategies are shown in \eqref{eq:magnitude}–\eqref{eq:entropy},
\begin{align}
g_k^{\text{magnitude}}(\mathbf{x}) &= \frac{\exp(\|\boldsymbol{\phi}_k(\mathbf{x})\|_2)}{\sum_{j=1}^{K} \exp(\|\boldsymbol{\phi}_j(\mathbf{x})\|_2)}, \label{eq:magnitude} \\
g_k^{\text{variance}}(\mathbf{x}) &= \frac{\exp(\text{Var}(\boldsymbol{\phi}_k(\mathbf{x})))}{\sum_{j=1}^{K} \exp(\text{Var}(\boldsymbol{\phi}_j(\mathbf{x})))} , \label{eq:variance} \\
g_k^{\text{entropy}}(\mathbf{x}) &= \frac{\exp(H_k(\mathbf{x}))}{\sum_{j=1}^{K} \exp(H_j(\mathbf{x}))}, \label{eq:entropy}
\end{align}
where $\boldsymbol{\phi}_k(\mathbf{x})$ represents extracted features and $H_k(\mathbf{x}) = -\sum_{b=1}^{20} p_{k,b}\log(p_{k,b}+\epsilon)$ is entropy from 20-bin histograms. Magnitude weighting emphasizes experts with stronger overall feature activations, variance weighting favors experts whose features display greater spread, and entropy weighting highlights experts with more uniform distributions. These methods differ in whether they capture signal strength, variability, or uncertainty, providing complementary perspectives for weighting experts.

\subsection{Learned Gating Network Architectures}
\subsubsection{Architecture-Specific Gating Formulations}
The CNN gating network applies multi-layer perceptron processing of FC1 features as in \eqref{eq:cnn1}–\eqref{eq:cnn3},
\begin{align}
\mathbf{h}_1 &= \text{ReLU}(\mathbf{W}_1 \overline{\boldsymbol{\phi}}^{\text{FC1}}(\mathbf{x})+\mathbf{b}_1), \quad \text{64 units}, \label{eq:cnn1} \\
\mathbf{h}_2 &= \text{ReLU}(\mathbf{W}_2 \mathbf{h}_1+\mathbf{b}_2), \quad \text{32 units}, \label{eq:cnn2} \\
\boldsymbol{z}(\mathbf{x}) &= \mathbf{W}_3 \mathbf{h}_2+\mathbf{b}_3, \quad 7 \text{ outputs}. \label{eq:cnn3}
\end{align}

The Mamba gating network incorporates multi-head attention in \eqref{eq:mamba1}–\eqref{eq:mamba3},
\begin{align}
\mathbf{h}_{\text{proj}} &= \mathbf{W}_{\text{proj}} \overline{\boldsymbol{\phi}}^{\text{pre-cls}}(\mathbf{x})+\mathbf{b}_{\text{proj}}, \label{eq:mamba1} \\
\mathbf{h}_{\text{attn}} &= \text{MultiHeadAttention}(\mathbf{h}_{\text{proj}},\mathbf{h}_{\text{proj}},\mathbf{h}_{\text{proj}}), \label{eq:mamba2} \\
\boldsymbol{z}(\mathbf{x}) &= \text{MLP}(\mathbf{h}_{\text{attn}}). \label{eq:mamba3}
\end{align}

The ViT gating network applies transformer-style processing with normalization as in \eqref{eq:vit1}–\eqref{eq:vit4},
\begin{align}
\mathbf{h}_{\text{proj}} &= \text{LayerNorm}(\mathbf{W}_{\text{proj}}\overline{\boldsymbol{\phi}}^{\text{CLS}}(\mathbf{x})+\mathbf{b}_{\text{proj}}), \label{eq:vit1} \\
\mathbf{h}_{\text{attn}} &= \text{LayerNorm}(\text{MultiHeadAttention}(\mathbf{h}_{\text{proj}})+\mathbf{h}_{\text{proj}}), \label{eq:vit2} \\
\mathbf{h}_{\text{mlp}} &= \text{LayerNorm}(\text{MLP}(\mathbf{h}_{\text{attn}})+\mathbf{h}_{\text{attn}}), \label{eq:vit3} \\
\boldsymbol{z}(\mathbf{x}) &= \mathbf{W}_{\text{head}}\mathbf{h}_{\text{mlp}}+\mathbf{b}_{\text{head}}. \label{eq:vit4}
\end{align}

The CNN gating network uses FC1 activations because they provide compact mid-level features, and a lightweight MLP can map them into expert weights efficiently. The Mamba gating network leverages state-space projection with multi-head attention to capture sequential structure and contextual dependencies in pre-classification features, which improves adaptive weighting. The ViT gating network applies transformer-style normalization and attention on the CLS token, enabling the gating module to model global context and assign expert weights in a more input-adaptive and context-aware manner.



\subsubsection{Gating Network Training Configuration}
Learned gating networks were trained with AdamW (learning rate $0.001$, weight decay $10^{-4}$), early stopping (patience 15), and gradient clipping (max norm 1.0). The loss function is given in \eqref{eq:gating_loss},
\begin{equation}
\mathcal{L}_{\text{gating}} = \mathcal{L}_{\text{BCE}}(\hat{y}_{\text{ensemble}},y) + \lambda \mathcal{L}_{\text{entropy}}(\mathbf{g}),
\label{eq:gating_loss} 
\end{equation}
with $\lambda=0.01$.

Final gating weights for learned networks are normalized as in \eqref{eq:learned_softmax}–\eqref{eq:learned_sigmoid},
\begin{align}
g_k^{\text{learned-softmax}}(\mathbf{x}) &= \frac{\exp(z_k(\mathbf{x}))}{\sum_{j=1}^{K}\exp(z_j(\mathbf{x}))}, \label{eq:learned_softmax} \\
g_k^{\text{learned-sigmoid}}(\mathbf{x}) &= \frac{\sigma(z_k(\mathbf{x}))}{\sum_{j=1}^{K}\sigma(z_j(\mathbf{x}))}. \label{eq:learned_sigmoid}
\end{align}

\subsection{End-to-End MoE Training Details}

\subsubsection{Hierarchical Weight Normalization}
Expert weights are normalized at lobe and lung levels as in \eqref{eq:w_lobes}–\eqref{eq:w_lungs},
\begin{align}
\mathbf{w}_{\text{lobes}} &= \frac{\mathbf{w}_{1:5}}{\sum_{i=1}^{5} w_i}, \label{eq:w_lobes} \\
\mathbf{w}_{\text{lungs}} &= \frac{\mathbf{w}_{6:7}}{\sum_{i=6}^{7} w_i}. \label{eq:w_lungs}
\end{align}

\subsubsection{Multi-Component Loss Function}
To ensure that the end-to-end model learns not only accurate predictions but also balanced and interpretable expert contributions, we design a multi-component loss function that combines standard classification loss with additional regularizers for entropy behavior, weight distribution, and expert diversity. The complete loss includes four components in \eqref{eq:loss_ce}–\eqref{eq:loss_div},
\begin{align}
\mathcal{L}_{\text{CE}} &= -\tfrac{1}{N}\sum_{i=1}^{N}\!\left[y_i\log(p_i)+(1-y_i)\log(1-p_i)\right], \label{eq:loss_ce} \\
\mathcal{L}_{\text{entropy}} &= -\sum_{k=1}^{K} w_k \log(w_k+\epsilon), \label{eq:loss_gating} \\
\mathcal{L}_{\text{weight}} &= \|\mathbf{w}-\tfrac{1}{K}\mathbf{1}\|_2^2, \label{eq:loss_weight} \\
\mathcal{L}_{\text{diversity}} &= \tfrac{1}{K^2}\sum_{i=1}^{K}\sum_{j \neq i}^{K}\!|\cos(\boldsymbol{\phi}_i,\boldsymbol{\phi}_j)|. \label{eq:loss_div}
\end{align}


\bibliographystyle{IEEEtran}
\bibliography{references}

\end{document}